# Persian Pronoun Resolution: Leveraging Neural Networks and Language Models

**Author information:**

1- Hassan Haji Mohammadi

   Islamic Azad University Tehran North Branch, Tehran, Iran

   h.hajimohammadi@iau-tnb.ac.ir

2- Alireza Talebpour
   *correspond author
   Shahid Beheshti University, Tehran, Iran

   Talebpour@sbu.ac.ir

3- Ahmad Mahmoudi Aznaveh
   Shahid Beheshti University, Tehran, Iran
   a_mahmoudi@sbu.ac.ir

4- Samaneh Yazdani
   Islamic Azad University Tehran North Branch, Tehran, Iran
   samaneh.yazdani@gmail.com

## Abstract

Coreference resolution, critical for identifying textual entities referencing the same entity, faces challenges in pronoun resolution, particularly identifying pronoun antecedents. Existing methods often treat pronoun resolution as a separate task from mention detection, potentially missing valuable information. This study proposes the first end-to-end neural network system for Persian pronoun resolution, leveraging pre-trained Transformer models like ParsBERT. Our system jointly optimizes both mention detection and antecedent linking, achieving a 3.37 F1 score improvement over the previous state-of-the-art system (which relied on rule-based and statistical methods) on the Mehr corpus. This significant improvement demonstrates the effectiveness of combining neural networks with linguistic models, potentially marking a significant advancement in Persian pronoun resolution and paving the way for further research in this under-explored area.

## 1. Introduction

Coreference resolution, a key task in natural language processing (NLP), entails identifying expressions in a text that refer to the same real-world entity or event. These expressions are then grouped together, as shown in Figure 1. This task is vital for various NLP applications such as question answering (Bhattacharjee, Haque, de Buy Wenniger, & Way, 2020), information extraction (Kriman & Ji, 2021), entity linking (Sil, Kundu, Florian, & Hamza, 2018), sentiment analysis (Mao & Li, 2021), and text summarization (Z. Liu, Shi, & Chen, 2021). Early coreference resolution systems heavily relied on handcrafted features, employing a pipeline approach that began with detecting expressions and clustering them (Clark & Manning, 2016a, 2016b; H. Lee et al., 2013; S. Wiseman, Rush, & Shieber, 2016; S. J. Wiseman, Rush, Shieber, & Weston, 2015). However, these systems faced limitations, notably errors introduced by syntactic parsers and language-specific feature tailoring (K. Lee, He, Lewis, & Zettlemoyer, 2017).

Anaphora resolution, a specialized subfield of coreference resolution, aims to identify the antecedents of anaphoric expressions within a given text. The primary goal of anaphora resolution is to locate the preceding mention (antecedent) of entities present in the text, which may include noun phrases, verbs, or pronouns. Anaphora resolution in languages like Persian poses greater challenges due to their distinct syntactic and morphological structures and the limited availability of preprocessing tools and diverse corpora. This research's core objective is to identify antecedents for pronouns within Persian text, a subset known as pronoun resolution, dedicated to ascertaining the antecedent of pronouns embedded within the text. Notably, this task presents challenges, particularly in pro-drop languages like Persian, where pronouns can exhibit ambiguity and have multiple references.

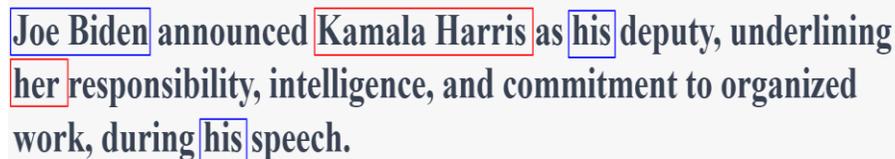

**Figure 1**. An illustration of coreference resolution is depicted in the image. The red and blue colors represent two separate clusters of mentions.

The advent of end-to-end (*e-2-e*) coreference resolution models in 2017 revolutionized the field of coreference resolution, paving the way for a new era of *e-2-e* systems that have outperformed traditional methods (K. Lee et al., 2017). These novel systems break away from the dependency on syntax parsers and handcrafted features, adopting a holistic approach that utilizes data-driven techniques and neural networks. Researchers have experimented with various techniques, including building global models and integrating advanced language models like BERT (Devlin, Chang, Lee, & Toutanova, 2018) and spanBERT (Joshi et al., 2020), to continuously enhance the efficiency and accuracy of coreference resolution (Bai, Zhang, Song, & Xu, 2021; Joshi et al., 2020; Joshi, Levy, Zettlemoyer, & Weld, 2019; Kantor & Globerson, 2019; Lai, Bui, & Kim, 2022; K. Lee, He, & Zettlemoyer, 2018; Wang, Shen, & Jin, 2021).

Coreference resolution systems can be broadly classified into four categories: feature-based models, multi-layer perceptron (MLP)/recurrent neural network (RNN) models, knowledge-based models, and transformer-based models (R. Liu, Mao, Luu, & Cambria, 2023). Feature-based systems rely on manually engineered features, which can be time-consuming and labor-intensive to construct. In contrast, neural network-based approaches offer end-to-end solutions that utilize neural networks to directly learn relationships between entities within text without explicitly incorporating external knowledge sources. Knowledge-based models, on the other hand, actively integrate external information and are constructed upon neural network architectures. Currently, transformer-based systems have garnered significant attention among researchers. These systems leverage pre-trained language models like BERT or spanBERT, demonstrating their potential to revolutionize the field of coreference resolution.

Despite boasting superior performance, end-to-end coreference models often suffer from high computational complexity, limiting their practical use. This has been particularly true for Persian, where coreference resolution has received comparatively less research attention. While traditional approaches relied heavily on manually crafted features, we present the first end-to-end Persian pronoun resolution system. This streamlined model leverages powerful transformer-based language models like BERT, eliminating the need for complex syntactic parsers or hand-crafted features.

By relying solely on gold mention data, our system pinpoints pronoun antecedents with remarkable accuracy. It offers several compelling merits, including enhanced interpretability, accelerated training and inference, reduced overfitting, improved generalization, simplified implementation, and reduced maintenance overhead. Additionally, the system augments mention representations by harnessing the power of various Persian pre-trained language models, such as ParsBERT (Farahani, Gharachorloo, Farahani, & Manthouri, 2021). The proposed model tackles the task of mention detection and pronoun resolution jointly. Notably, the proposed system surpasses the performance of all prior systems, demonstrating its remarkable efficacy. To evaluate our system's efficacy, we conducted extensive experiments on the Mehr Persian coreference corpus (Haji Mohammadi, Talebpour, Mahmoudi Aznaveh, & Yazdani, 2023). Our results establish a new benchmark for Persian pronoun resolution, surpassing all previous systems. Specifically, our system achieves a precision of 68.91%, a recall of 67.35%, and an F1 score of 68.12%, demonstrating its substantial improvement over prior approaches.

The following sections delve deeper into our research. Section 2 commences with an overview of related work in the domain of knowledge-poor languages. Section 3 then unveils our proposed model, meticulously detailing its architecture and components. Next, Section 4 meticulously describes the conducted experiments, encompassing the dataset, evaluation methodology, and employed baseline systems. Section 5 presents the study's findings, including comprehensive discussions, and a thorough examination of the results. Finally, Section 6 concludes the paper, summarizing key insights and highlighting the significance of our work.

## 2. Related Work

Coreference resolution has been extensively studied in knowledge-rich languages like English, with numerous comprehensive reviews providing valuable insights (Lata, Singh, & Dutta, 2022; R. Liu et al., 2023; Stylianou & Vlahavas, 2021; Sukthanker, Poria, Cambria, & Thirunavukarasu, 2020). These reviews categorize the research landscape into four main categories: feature-based methods (Denis & Baldridge, 2008; Durrett & Klein, 2014; H. Lee et al., 2013; Soon, Ng, & Lim, 2001), multilayer perceptron/recurrent neural networks (Clark & Manning, 2016a, 2016b; K. Lee et al., 2017; S. Wiseman et al., 2016; Xu & Choi, 2020), knowledge-based systems (Aralikatte et al., 2019), and transformer-based systems (Joshi et al., 2020; Joshi et al., 2019; Khosla & Rose, 2020; Lai et al., 2022; Wu, Wang, Yuan, Wu, & Li, 2020; Xia, Sedoc, & Van Durme, 2020).

However, research in knowledge-poor languages like Persian has been relatively limited in recent years. This section explores the work conducted in similar low-resource language settings, namely Arabic and Turkish. Subsequently, we embark on an in-depth examination of Persian pronoun resolution systems, highlighting the unique challenges and advancements in this linguistic domain.

In Turkish, a work has been conducted by (Demir, 2023), , introducing two neural models utilizing the mention-ranking approach and evaluated on a Turkish coreference dataset. The first model employs handcrafted features to measure the similarity between mention pairs, while the second adopts an end-to-end coreference scoring approach, effectively combining embeddings and attention mechanisms to identify antecedents.

Additionally, in the Indonesian context, Auliarachman and Purwarianti (Auliarachman & Purwarianti, 2019) present a coreference resolution system that leverages neural models, incorporating features such as mention words, contextual information, and mention pair relationships. Their methodology involves concatenating feature representations, employing Convolutional Neural Networks (CNNs) and Fully Connected Networks (FCNs), and implementing a greedy best-first algorithm for mention clustering. These advancements represent significant strides in coreference resolution for languages like Turkish and Indonesian, demonstrating the effectiveness of neural approaches in these low-resource settings.

Aloraini and Massimo (Aloraini & Poesio, 2020) present a novel BERT-based model for Arabic zero pronoun resolution. It outperforms previous methods, including BERT feature extraction and fine-tuning, while also pinpointing the most suitable layer in BERT for this task. Murayshid et al.'s approach (Murayshid, Benhidour, & Kerrache, 2023) employs a synergistic combination of a pre-trained Arabic BERT encoder and a Bi-LSTM decoder to effectively assess the likelihood of tokens being accurate antecedents. During training, they minimize binary cross-entropy loss to optimize the alignment between predicted scores and actual values, leading to enhanced resolution accuracy.

Early endeavors in utilizing neural networks for anaphora resolution primarily centered on Japanese and Chinese languages, particularly tackling zero pronoun resolution, where pronouns are omitted. A pioneering study (Yin, Zhang, Zhang, Liu, & Wang, 2018), introduced an RNN-based approach that incorporated attention mechanisms to address the complexities associated with zero-pronoun resolution. These advancements in neural network-based approaches have made substantial contributions to anaphora resolution, particularly in languages like Arabic, Japanese, and Chinese, where handling zero pronouns poses unique challenges.

Persian coreference and pronoun resolution, despite its growing importance, has a relatively short history. Most prior systems in this domain concentrated primarily on coreference resolution. These pioneering Persian coreference and pronoun resolution systems were developed using limited corpora, lacking the comprehensive end-to-end capabilities necessary to generate complete coreference resolution chains. Evaluating these systems comparatively proved time-consuming due to imprecise assessment methods and reliance on outputs generated by data mining tools like Weka. Persian coreference and pronoun resolution has faced various limitations and constraints, hindering the development of comprehensive and accurate systems.

In 2009, Sadat Mousavi and Ghassem-Sani (Moosavi & Ghassem-Sani, 2009) pioneered the field of Persian pronoun resolution by introducing the first Persian pronoun resolution system. This system employed algorithms like decision trees and maximum entropy, leveraging a ranking approach on the PCAC-2008 corpus. Subsequently, in 2016, Hajimohammadi et al. (Hajimohammadi, Talebpour, & Mahmoudi Aznaveh, 2016) achieved a remarkable 20% performance improvement by incorporating Persian-language specific features into their pronoun resolution system. Their approach demonstrated the importance of tailoring features to the linguistic nuances of the target language.

Nourbaksh and Bahrani (Nourbakhsh & Bahrani, 2017) further enhanced the system by integrating a wider range of features and refining their sampling method, leading to improved performance. Their work highlighted the benefits of expanding feature representations and optimizing sampling techniques. Fallahi and Shamsfard (Fallahi & Shamsfard, 2011), developed a rule-based pronoun resolution system, utilizing handcrafted rules to analyze three-sentence windows. This approach demonstrated the potential of rule-based systems, though it faced limitations in handling complex linguistic structures.

Nazaridoust et al. (Nazaridoust, Bidgoli, & Nazaridoust, 2013) introduced the first Persian coreference resolution system on the Lotus corpus. Their system employed feature construction techniques and classifiers such as decision trees, neural networks, and SVMs, marking a significant advancement in the field. It is crucial to note that these systems primarily provided statistical analyses without generating conclusive coreference chains, highlighting a key limitation in their functionality.

Recent advancements have propelled the development of contemporary Persian systems that establish definitive coreference chains on current corpora. Rahimi and HosseinNejad (Rahimi & HosseinNejad, 2020) spearheaded this progress by introducing a coreference system on a standardized corpus, enabling the creation of conclusive coreference chains. Sahlani et al. (Sahlani, Hourali, & Minaei-Bidgoli, 2020a, 2020b) furthered these advancements by introducing graph-based and neural network approaches that seamlessly integrate both hand-extracted and embedding features, significantly enhancing coreference resolution efficiency. Haji Mohammadi et al. (Mohammadi, Talebpour, Aznaveh, & Yazdani, 2022) brought forth a hybrid entity-based pronoun resolution system that outperformed earlier methods on the Mehr and RCDAT corpora (Rahimi & HosseinNejad, 2020) , further solidifying the progress made in this domain. These modern systems offer improved accuracy and comprehensive coreference resolution through standardized corpora and advanced methodologies.

Despite these remarkable strides, it is important to acknowledge that these systems have continued to rely on handcrafted features in constructing their feature vectors. In this article, we embark on a journey by presenting an end-to-end model derived from pre-trained language models, dismantling the reliance on handcrafted features in a Persian pronoun resolution system and marking a significant contribution in this field.

## 3. Proposed Method

The system presented in this article adopts the end-to-end (*e-2-e*) architecture as its core framework, marking a significant departure from previously developed Persian systems that relied heavily on labor-intensive handcrafted features and syntactic parser. Unlike English systems, which often incorporate language models and attention mechanisms, Persian systems have rarely ventured into these advanced architectural territories. A hallmark of the *e-2-e* model lies in its conscious avoidance of handcrafted mention detection and its embrace of a data-driven approach.

The proposed system employs a multi-stage process for pronoun resolution, leveraging the power of transformer-based language models. The initial stage involves generating token representations using a pre-trained BERT model. Subsequently, in-text spans with a maximum length of $L$ are extracted. The antecedent pruning approach effectively identifies high-scoring spans, which are then passed to the subsequent stage to establish a distribution between spans and their corresponding candidate antecedents. This distribution underpins the construction of final links, enabling the evaluation of pronoun resolution accuracy. The ensuing sections delve into the intricacies of each stage.

### 3-1 Transformer-based Models

Sequence-to-sequence models, as described in (Sutskever, Vinyals, & Le, 2014), have revolutionized various domains, including machine translation. These neural networks excel at mapping input sequences to output sequences, irrespective of their length. Prior to the advent of transformer architectures, sequence-to-sequence models primarily relied on recurrent neural networks (RNNs) and convolutional neural networks (CNNs). For instance, in the context of Arabic pronoun resolution, Murayshid et al. (Murayshid et al., 2023) employed Bi-LSTM in their sequence-to-sequence system.

However, the landscape shifted significantly with the introduction of the Transformer architecture by Vaswani et al. (Vaswani et al., 2017) in 2017. This innovative model, relying on the attention mechanism, departed from using RNN and CNN structures. The transformer architecture offers several advantages over its predecessors, including accelerated processing, enhanced parallelism, complete connectivity, and the integration of self-attention mechanisms. These advancements have significantly impacted the field of natural language processing, particularly in coreference resolution systems.

BERT (Bidirectional Encoder Representations from Transformers) (Kenton & Toutanova, 2019) is a revolutionary language model that leverages the Transformer architecture to capture the contextual meaning of text without relying on a fixed word order. The integration of BERT into coreference resolution has significantly enhanced the performance of these systems by providing a comprehensive understanding of the surrounding context.

Joshi et al. (Joshi et al., 2019) pioneered the application of BERT to coreference resolution by introducing a model based on the *c-2-f* model (K. Lee et al., 2018). Their model achieved remarkable performance gains by replacing the conventional LSTM encoder with BERT. In a subsequent work, Joshi et al. (Joshi et al., 2020) further improved the performance of their BERT-based coreference resolution system by substituting BERT with SpanBERT, a variant of BERT that employs an alternative masking approach during training. This innovative masking strategy yielded substantial improvements in the system's efficiency and overall performance.

A recent study by Hou et al. (Hou et al., 2024) challenges the conventional wisdom that Higher-Order Information (HOI) is beneficial for coreference resolution. They argue that HOI may actually have marginal or even negative effects on coreference resolution performance. To address this concern, the researchers propose two novel methods: Less-Anisotropic Internal Representations (LAIR) and Data Augmentation with Document Synthesis and Mention Swap (DSMS). These approaches are designed to learn less-anisotropic span embeddings, which are shown to improve the performance of coreference resolution tasks.

### 3-2 Notation and Problem Formulation

The development of coreference resolution systems typically follows a two-stage procedure: mention detection and mention linking. This section delves into the development of Persian pronoun resolution within the *e-2-e* architecture. Let's represent the current document under consideration as $D = (n_1, n_2, ..., n_T)$, where $T$ represents the number of text tokens. Assume that the number of spans in a given text is $N = T(T+1)/2$. The foundational *e-2-e* model, commences by identifying all spans within the text, followed by mention detection on the spans with the highest scores. In our research, we employ the span identification approach proposed by (Lai et al., 2022), which leverages a hierarchical neural network to identify all potential antecedent candidates for pronouns in a text.

In an end-to-end coreference resolution system, each span, denoted as *i*, is assigned a unique antecedent, $y_i$. A potential span candidate for each span *i* is chosen from the set $Y_i = [1, ..., i-1, \varepsilon]$. Here, $\varepsilon$ signifies that either (1) span *i* does not represent an entity mention or (2) span *i* serves as the initial mention of a new cluster. In the domain of coreference resolution, aimed at constructing clusters of co-referent entities, these entity clusters evolve through a transitive relationship. This process aligns with the overarching objective of assembling coherent and interconnected references within the text. Our primary objective is to learn a distribution $P(y_i)$ over antecedents for each span:

$$P(yi) = \frac{e^{s(i,y_i)}}{\sum_{y' \in Y(i)} e^{s(i,y')}} \tag{1}$$

In this context, the pairwise score s(i,j) assesses the likelihood of span *j* being an antecedent of span *i*. The formula incorporates three key factors: (1) $s_m(i)$: A binary indicator determining whether span *i* is a mention or a pronoun, $s_m(j)$: A binary indicator determining whether span *j* is a mention, and $s_a(i, j)$: a factor indicating the presence of a coreference link between span *i* and span *j*:

$$s(i,j) = \begin{cases} Sm(i) + Sm(j) + Sa(i,j), & j \neq \varepsilon \\ 0 & j = \varepsilon \end{cases} \tag{2}$$

We also used a combination of two methods from Lee et al. (K. Lee et al., 2018) to capture global features: coarse-to-fine antecedent pruning and representation refining. Coarse-to-fine antecedent pruning: Gradually eliminates unlikely antecedent candidates using hierarchical clustering based on pairwise similarities. Representation refining: Iteratively updates span representations by incorporating information from antecedent candidates, enhancing their suitability for accurate linking.

### 3-3 Representation Layer

Each token within the document *D* is represented as a d-dimensional vector, referred to as *vi* ∈ $R^d$. This d-dimensional vector functions as a contextualized representation for each input token and is produced by a transformer-based encoder, such as ParsBERT (Farahani et al., 2021). We choose ParsBERT over other pre-trained models due to its effectiveness in capturing Persian language nuances and its alignment with our *e-2-e* framework. This paper leverages the full potential of BERT by employing both feature extraction and fine-tuning. The pretrained language models are limited to sequences of up to 512 tokens. To handle longer documents, overlapping segments are used (Joshi et al., 2019). These segments are then independently passed through the BERT encoder to obtain the final representation for all text tokens.

For each span *k* in a text, the span representation is composed of four components: the embedding of the first token, the embedding of the last token, a weighted combination of all tokens, and *φ(i)* represents a feature vector used to determine whether the given span functions as a pronoun. The span representation *gi* is defined as follows:

$$g_i = [x^*_{START(k)}; x^*_{END(k)}; \hat{x}_k; \varphi(i)] \tag{1}$$

In this equation, $x^*_{START(k)}$ represents the embedding vector of the first token in the span, $x^*_{END(k)}$ represents the embedding vector of the last token in the span and $\hat{x}_k$ is the soft head word vector calculated using the attention mechanism (Bahdanau, Cho, & Bengio, 2015) as follows:

$$\alpha_t = w_\alpha . FFNN_\alpha(x^*_t) \tag{2}$$

$$\delta_{k,t} = \frac{\exp(\alpha_t)}{\sum_{i=start(k)}^{end(k)} \exp(\alpha_t)} \tag{3}$$

$$\hat{x}_k = \sum_{i=start(k)}^{end(k)} \delta_{k,i} . x_i \tag{4}$$

With the head attention mechanism, each mention undergoes processing through a feed-forward neural network (FFNN), and each token within the mention is assigned a score. The weighted sum of $\hat{x}_k$ is then concatenated with $g_i$. This approach bears similarities to the *e-2-e* coreference resolution method, except that it does not incorporate additional features.

### 3-3-2 Model

In the first step, all in-text spans are detected, up to a maximum length of *L*. These spans are then input to a feedforward neural network to calculate the scoring functions:

$$S_m(i) = U_m^T . FFNN_m(g_i) \tag{5}$$

$$S_a(i,j) = U_a^T . FFNN_a([g_i, g_j, g_i \circ g_j]) \tag{6}$$

where $o$ in this equation denotes element-wise multiplication. When a span is not a mention or doesn't refer to any real-world entity, its coreference score with a dummy antecedent is zero. This means that $S_a(i, \varepsilon) = 0$. Lee et al. (2017) highlight the computational burden of constructing mention pairs when considering all candidate spans, leading to an $O(T^4)$ complexity. To tackle this challenge, we examine spans up to $L$ words.

During both training, span pruning is used to efficiently target the most promising candidates. Only the top $\lambda N$ spans, where $N$ represents the total number of spans identified, are retained. This selection is based on their "mention scores," which indicate their relevance and informativeness. Lower-scoring spans are discarded to streamline processing and reduce computational overhead. In this paper, the value of $\lambda$ is set to 0.4 during the development phase.

In the joint coreference resolution model proposed by Lee et al. (2017), the mention detector is not pre-trained separately, leading to random initialization of spans in the joint model, which can hinder training efficiency. To address this issue, we adopt the approach introduced by Zhang et al. (2018) (Zhang, dos Santos, Yasunaga, Xiang, & Radev, 2018). In this method, the mention detector undergoes initial pre-training using gold coreference chains from the Mehr corpus, enabling direct optimization of the mention detector:

$$L_{\text{detect}}(i) = y_i \log \hat{y}_i + (1 - y_i) \log(1 - \hat{y}_i) \tag{7}$$

In this equation, $\hat{y}_i = sigmoid(Sm(i))$, and $y_i$ equals one when spans is an exact match to a gold mention. The final loss function integrates both the mention detection loss and the clustering loss:

$$L = \lambda_{detect} \sum_{i \in S} Ldetect(i) + \sum_{i \in U} Lcluster(i) \tag{8}$$

$S$ denotes the set of all spans, and $U$ represents the set of unpruned spans after pruning. The parameter $\lambda_{detect}$ regulates the weightage assigned to the two distinct losses in the model. The clustering loss is defined as follows:

$$L_{cluster}(i) = -\log \sum_{\hat{y} \in (Yi) \cap GOLD(i)} P(\hat{y}) \tag{9}$$

Where $Y_i$ represents the set of potential antecedent spans for $i$, and GOLD(i) denotes the set of gold antecedent spans for span $i$. GOLD(i) = $\varepsilon$ if span $i$ does not belong to any coreference chains or if all its gold antecedents are pruned. The proposed architecture is depicted in Figure 2.

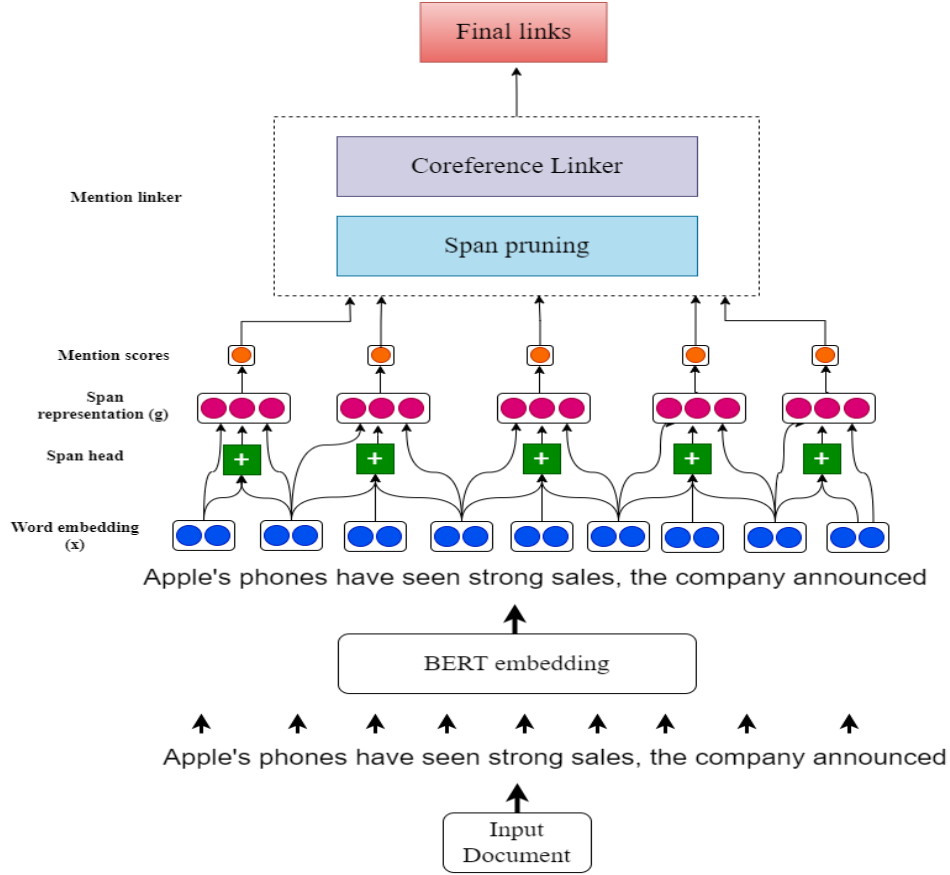

Figure 2. The overall proposed architecture of the Persian Pronoun Resolution system. All spans up to size L are considered in the final implementation, although only a few are shown in the figure for illustrative purposes.

## 4. Experiments

This section outlines the core methodology of the research. It describes the key steps in the experiments, including the chosen dataset, evaluation methods, hyperparameter tuning, and exploration of context-dependent word embeddings and comparison systems. The use of the Mehr corpus and the focus on the F1 score for evaluation are discussed, followed by an examination of hyperparameter tuning to optimize model performance. Finally, experiments exploring context-dependent word embeddings and comparing the model to other pronoun resolution systems are presented.

**4-1 Dataset and Performance Measures**

In our research on pronoun resolution systems, we leverage the Mehr corpus (Haji Mohammadi et al., 2023) due to its unique strengths. Unlike corpora like RCDAT, Mehr provides fine-grained annotations for singleton pronouns, which are crucial for our model's training. Spanning diverse genres such as politics, sports, and economics, the corpus offers a realistic representation of language use in various contexts. Notably, its Test section comprises 13,000 tokens, mirroring the size of the MUC corpus (Hirschman, 1997) Test section, enabling a direct comparison of our model's performance on tasks with similar complexity.

The Mehr corpus utilizes the CoNLL standard (Pradhan, Moschitti, Xue, Uryupina, & Zhang, 2012) and WebAnno tool (Yimam, Gurevych, de Castilho, & Biemann, 2013) for annotation. Notably, gold labels are assigned only to mentions involved in coreference relationships, alongside additional identification of singleton pronouns. This includes explicit delineation of various pronoun types, such as personal, reflexive, and demonstrative. Furthermore, natural language processing tools automatically generated some labels, including 44,960 nominal phrases and 11,175 nominal entities. Importantly, 19,833 mentions are designated as gold mentions participating in coreference relationships In Table 1, the specifications of the Mehr corpus are provided.

Table 1. Statistical Information of the Mehr Corpus.

|  | Document counts | Sentence counts | Token counts |
| --- | --- | --- | --- |
| **Train and development** | 357 | 3189 | 158047 |
| **Test** | 43 | 286 | 13214 |
| **Total** | 400 | 3475 | 171261 |

The critical role of accurate evaluation metrics cannot be understated, particularly in pronoun resolution systems where performance directly influences applications like natural language understanding and machine translation. We face a fundamental challenge in our research: the inherent imbalance in our carefully curated dataset. Traditional metrics, like accuracy, often require adaptation to effectively assess systems under such conditions. Imbalanced datasets can mask a system's true capabilities, potentially leading to high performance on the majority class while neglecting the crucial, but less frequent, minority class.

To mitigate the challenges of an imbalanced dataset, we adopt the F1-score as our primary evaluation metric. This versatile metric considers both true positives and handles false positives and negatives, balancing precision and recall. This combined analysis provides a more nuanced and robust assessment of our pronoun resolution system's performance, especially in the face of imbalanced data.

### 4-2 Hyper Parameters

A cornerstone of any robust neural system lies in its innovative architecture and meticulously calibrated hyperparameters. Recognizing the critical role of hyperparameter optimization in crafting a high-performing, rigorously tested Persian pronoun resolution system, we have meticulously selected and fine-tuned these parameters to ensure exceptional resilience and adaptability across diverse conditions. Leveraging insights from Lai et al. (Lai et al., 2022) system while considering the unique nuances of the target language and corpus, our proposed system delivers both efficiency and accuracy in Persian language pronoun resolution.

To address computational limitations arising from the small number of mention sizes in our corpus, the maximum span size was effectively reduced from 30 to 10. Additionally, to strike a balance between accurate pronoun detection and computational efficiency, the top span ratio ($\lambda$) was set to 0.4 and the weight of detection loss ($\lambda$-detect) was adjusted to 0.2. Furthermore, aiming for optimal performance, the number of FFNN nodes was set to 1200. Finally, to ensure efficient training convergence, the model underwent 100 epochs using the Adam algorithm (Kingma & Ba, 2014), leveraging regularization techniques like early stopping, gradient clipping,

### 4-3 More Experiments

To comprehensively analyze our model's capabilities and adaptability, we conducted a series of experiments exploring various dimensions. These evaluations aimed to deepen our understanding of its performance under diverse operating conditions.

To enrich our understanding of context-dependent word representations in Persian, we explored utilizing embeddings from various language models beyond ParsBERT v1.0. Each word meaning can change based on the surrounding words, and these models capture those nuances through different "embeddings." Our study incorporated embeddings from ParsBERT v1.0 (bert-base-parsbert-uncased) as a baseline, then broadened its scope by including a diverse range of Persian language models and their respective versions. This allowed us to delve deeper into the richness and complexity of context-dependent representations within the Persian language. These models include:

- BERT v3.0 Model (HooshvareLab/bert-fa-zwnj-base): This model from HooshvareLab provides unique insights into context-dependent word embeddings designed explicitly for the Persian language.
- DistilBERT v3.0 Model (HooshvareLab/distilbert-fa-zwnj-base): Leveraging the DistilBERT architecture, this model refines understanding context-dependent embeddings in Persian.
- ALBERT v3.0 Model (HooshvareLab/albert-fa-zwnj-base-v2): HooshvareLab's ALBERT model introduces its take on context-dependent embeddings in Persian text analysis.
- ROBERTA v3.0 Model (HooshvareLab/roberta-fa-zwnj-base): The ROBERTA model, also from HooshvareLab, contributes to exploring context-dependent embeddings in Persian text analysis.

### 4-4 competing systems

In this section, we introduce the competing systems that serve as benchmarks for evaluating the performance of our pronoun resolution model. These systems encompass a range of approaches, including both Persian and English coreference resolution models. To comprehensively evaluate existing approaches, we re-implemented several pronoun resolution systems, including both Persian and non-Persian models. This facilitated an in-depth analysis of their strengths and weaknesses. Based on these insights, we carefully configured our own pronoun resolution system, tailoring it to the specific characteristics of the Persian language. Notably, when implementing non-Persian models, we carefully adapted them to the Persian context, potentially adding or removing features as needed to optimize their performance.

Among these, Hajimohammadi et al.'s (Mohammadi et al., 2022) hybrid approach specifically targets Persian pronoun resolution. Their model merges rule-based sieves (seven, highly accurate ones) with a machine-learning sieve tailored for pronouns. This novel combination leverages linguistic information to identify potential antecedents and refines pronoun-antecedent linking via a random forest classifier. Notably, they also introduced the "Mehr" corpus, significantly expanding available Persian coreference resources. This hybrid approach has demonstrated substantial performance improvements over prior Persian systems.

Sahlani et al. (Sahlani et al., 2020a) explored a fully connected neural network for Persian coreference resolution. Their model excelled in both feature extraction and mention pair classification by effectively combining handcrafted, word embedding, and semantic features. A deep neural network carefully assessed the validity of mention pairs, while a hierarchical accumulative clustering method identified coreference pairs. This approach achieved a noteworthy F-score of 64.54% on the Uppsala dataset, surpassing contemporary methods.

Lee et al. (K. Lee et al., 2017) made a groundbreaking contribution by introducing the first end-to-end coreference resolution model. This innovative model eliminated the need for syntactic parsers or hand-crafted mention detectors by considering all document spans as potential mentions. It relied on span embeddings that combined context-dependent boundary representations with a head-finding attention mechanism. Trained to maximize the likelihood of gold antecedent spans from coreference clusters, it achieved state-of-the-art performance without requiring external resources. Notably, this model serves as a valuable baseline for our proposed system. Additionally, it has been further enhanced by incorporating BERT embeddings, along with fine-tuned hyperparameters.

Murayshid et al. (Murayshid et al., 2023) propose a novel pronoun resolution system for Arabic, leveraging the power of BERT for encoding and a bi-directional LSTM (BI-LSTM) network (Graves & Schmidhuber, 2005) for decoding. Their model cleverly assigns a probability score to each token in the input sequence, effectively scoring on whether it belongs to the intended pronoun based on the generated binary output.

Lai et al. (Lai et al., 2022) revolutionized coreference resolution by introducing a remarkably simple yet effective system that eschews global features. It surpasses the previous state-of-the-art model (K. Lee et al., 2017) by a significant margin. Their clever approach lies in their scoring function, which simultaneously optimizes both mention detection and mention clustering in a unified step.

## 5. Results and Discussion

This section dives into the performance of our proposed pronoun resolution model, scrutinized through established metrics like precision, recall, and *F1* score. These multifaceted measurements provide a comprehensive view of the model's effectiveness in identifying and linking pronouns accurately. By delving into these rigorous assessments, we aim to unveil both the strengths and potential areas for improvement within our model, offering a nuanced and insightful perspective on its efficacy and reliability.

Our analysis, as presented in Table 2 , reveals the clear dominance of BERT-based configurations, particularly *bert-fa-zwnj-base*. This model demonstrates a significant advantage over traditional embedding methods like GLOVE and FASTTEXT, as well as contextualized embeddings like ROBERTA and DISTILBERT. Notably, *bert-fa-zwnj-base* achieved a remarkable F1 score of 68.12%, outperforming GLOVE (64.7%), FASTTEXT (64.29%), ROBERTA (65.78%), and DISTILBERT (60.78%). This resounding success underscores the effectiveness of BERT embeddings in capturing the nuances of Persian pronoun resolution and represents a substantial improvement over previous approaches.

Our model achieves performance gains by deviating from previous Persian approaches that rely on manual mention detection and handcrafted features. Instead, it leverages the power of pre-trained language models and attention mechanisms. Pre-trained language representations equip the model with rich contextual understanding, enabling it to adapt effectively to various linguistic contexts. Furthermore, attention mechanisms allow the model to pinpoint crucial details, leading to more accurate pronoun resolution.

Our pronoun resolution system surpasses existing approaches in several key ways. Compared to Lee et al. (K. Lee et al., 2017) , we leverage global features, integrate pronoun-specific features, and utilize the powerful BERT architecture. This not only enhances mention detection but also optimizes mention linking simultaneously, leading to a significant F1 score improvement of 7.07 points. While our system shows comparable performance to Lai et al. (Lai et al., 2022), the inclusion of global architecture and pronoun-specific features offers a slight edge. Finally, our well-designed architecture outperforms Murayshid encoder-decoder system by 1 point in F1 score, demonstrating its effectiveness.

While our model shows notable improvements, there is still room for further development. We identified two main types of errors:

- Coreference resolution within complex sentences: The model sometimes struggles with pronoun resolution in sentences with multiple potential antecedents or complex grammatical structures. This suggests further exploration of deeper contextual understanding and attention mechanisms.
- Pronoun disambiguation: Occasionally, the model misidentifies the correct referent for a pronoun, especially when pronouns have multiple potential antecedents with similar semantic roles. This points towards the need for incorporating more sophisticated reasoning and disambiguation techniques.

Our proposed model leverages BERT-based embeddings and a meticulously designed architecture to surpass not only traditional embedding methods but also previous Persian pronoun resolution systems. This superiority stems from the careful integration of pre-trained language models and attention mechanisms, as reflected in the significant improvement in F1 scores. The model's adaptability, demonstrated by its outperformance compared to English baselines, solidifies its position as a cutting-edge and versatile solution for Persian pronoun resolution. This advancement holds promise for broader natural language processing tasks and establishes the model as a benchmark for cross-linguistic pronoun resolution.

Table 2. Results on the test set on the Mehr corpus.

| Model | Precision | Recall | F1 Score |
|---|---|---|---|
| **Proposed Model** | | | |
| Persian e2e (fastText-200) | 64.29 | 63.95 | 64.12 |
| Persian e2e (fastText-300) | 64.45 | 64.15 | 64.29 |
| Persian e2e (glove-300) | 65.3 | 64.10 | 64.7 |
| Persian e2e (bert-base-parsbert-uncased) – fine tuning | 67.10 | 65.98 | 66.54 |
| Persian e2e (bert-fa-zwnj-base) – fine tuning | **68.91** | **67.35** | **68.12** |
| Persian e2e (bert-fa-zwnj-base) – feature extraction | 66.13 | 64.53 | 65.32 |
| Persian e2e (distilbert-fa-zwnj-base) – fine tuning | 61.04 | 60.52 | 60.78 |
| Persian e2e (albert-fa-zwnj-base-v2) – fine tuning | 66.01 | 64.70 | 65.35 |
| Persian e2e (roberta-fa-zwnj-base) – fine tuning | 66.39 | 65.18 | 65.78 |
| **Persian systems** | | | |
| (Sahlani et al., 2020a) | 60.47 | 59.88 | 60.47 |
| (Mohammadi et al., 2022) | 65.4 | 64.11 | 64.75 |
| **Other Baseline** | | | |
| (Murayshid et al., 2023) | 68.02 | 66.22 | 67.11 |
| (K. Lee et al., 2017) | 62.16 | 59.96 | 61.04 |
| + BERT | 64.78 | 63.04 | 63.9 |
| + BERT + hyper parameter tuning | 65.25 | 63.67 | 64.45 |
| (Lai et al., 2022) | 68.35 | 66.77 | 67.55 |

## 6. Conclusion

This study presents the development and evaluation of a state-of-the-art Persian pronoun resolution system, tackling the inherent challenges of imbalanced datasets and linguistic complexities in Persian. Utilizing the meticulously annotated Mehr corpus, we conducted extensive experiments, hyper-parameter optimization, and comparative analyses to deliver a comprehensive evaluation of our proposed model.

The F1 score proved indispensable as our primary metric, particularly for imbalanced datasets. Balancing precision and recall, it offered a nuanced understanding of our system's performance. The Mehr corpus, meticulously annotated with singleton pronouns and adhering to the CoNLL standard, enabled meaningful comparisons with established benchmarks like the MUC corpus. Hyper-parameter optimization was crucial in crafting a robust model. Our meticulously designed architecture, featuring carefully calibrated layers, dropout rates, and weight initialization, demonstrated resilience to overfitting and adaptability to diverse data scenarios. Furthermore, incorporating various Persian language models (BERT, DistilBERT, ALBERT, ROBERTA) yielded valuable insights into context-dependent embeddings, enriching our analysis.

Our proposed model outperformed previous Persian pronoun resolution systems by a significant margin, as demonstrated in comparisons with adapted and extended English baselines. This comparison effectively highlights the adaptability and effectiveness of our architecture. Notably, configurations that leverage BERT, particularly the bert-fa-zwnj-base model, achieved exceptional results, significantly surpassing traditional embedding methods like GLOVE and FASTTEXT and even outperforming contextualized embeddings like ROBERTA and DISTILBERT. On the Mehr corpus test set, bert-fa-zwnj-base achieved a remarkable F1 score of 68.12%, not only surpassing previous Persian pronoun resolution models but also overshadowing both traditional and contextualized embedding methods. Furthermore, the model's ability to perform well across linguistic boundaries, as evidenced by its comparison to the English baselines, solidifies its position as a versatile and effective solution for pronoun resolution.

This study introduces a robust Persian pronoun resolution model harnessing advanced language models and attention mechanisms. By meticulously addressing challenges like imbalanced datasets through hyper-parameter optimization and comparative analyses, we contribute significantly to the advancement of this field. Our proposed model not only outperforms existing systems but also holds promise as a benchmark for cross-linguistic pronoun resolution, paving the way for further research in natural language understanding and machine translation within the Persian language domain.

Looking ahead, we envision several exciting future research directions. We plan to investigate deeper contextual understanding and attention mechanisms for handling complex sentences. Additionally, incorporating more sophisticated linguistic features and reasoning techniques could further improve pronoun disambiguation. Furthermore, exploring domain adaptation techniques holds promise for enhancing generalizability across various domains. We also aim to analyze the model's performance on specific linguistic phenomena relevant to Persian and apply it to real-world NLP tasks like machine translation and question answering. By pursuing these directions, we believe this work can contribute significantly to the advancement of Persian pronoun resolution and NLP research, ultimately leading to improved capabilities for understanding and processing Persian language.

## Declaration of competing interest

The authors declare that they have no known competing financial interests or personal relationships that could have appeared to influence the work reported in this paper.

## Data availability

Data will be made available on request.

## Funding

No funding was received to assist with the preparation of this manuscript.

## Conflict of interests

The authors have no conflicts of interest to declare.

# References

Aloraini, A., & Poesio, M. (2020). *Cross-lingual zero pronoun resolution.* Paper presented at the Proceedings of the Twelfth Language Resources and Evaluation Conference.

Aralikatte, R., Lent, H., González, A. V., Herschcovich, D., Qiu, C., Sandholm, A., . . . Søgaard, A. (2019). *Rewarding Coreference Resolvers for Being Consistent with World Knowledge.* Paper presented at the Proceedings of the 2019 Conference on Empirical Methods in Natural Language Processing and the 9th International Joint Conference on Natural Language Processing (EMNLP-IJCNLP).

Auliarachman, T., & Purwarianti, A. (2019). *Coreference Resolution System for Indonesian Text with Mention Pair Method and Singleton Exclusion using Convolutional Neural Network.* Paper presented at the 2019 International Conference of Advanced Informatics: Concepts, Theory and Applications (ICAICTA).

Bahdanau, D., Cho, K. H., & Bengio, Y. (2015). *Neural machine translation by jointly learning to align and translate.* Paper presented at the 3rd International Conference on Learning Representations, ICLR 2015.

Bai, J., Zhang, H., Song, Y., & Xu, K. (2021). *Joint Coreference Resolution and Character Linking for Multiparty Conversation.* Paper presented at the Proceedings of the 16th Conference of the European Chapter of the Association for Computational Linguistics: Main Volume.

Bhattacharjee, S., Haque, R., de Buy Wenniger, G. M., & Way, A. (2020). *Investigating query expansion and coreference resolution in question answering on BERT.* Paper presented at the International conference on applications of natural language to information systems.

Clark, K., & Manning, C. D. (2016a). *Deep Reinforcement Learning for Mention-Ranking Coreference Models.* Paper presented at the Proceedings of the 2016 Conference on Empirical Methods in Natural Language Processing.

Clark, K., & Manning, C. D. (2016b). *Improving Coreference Resolution by Learning Entity-Level Distributed Representations.* Paper presented at the Proceedings of the 54th Annual Meeting of the Association for Computational Linguistics (Volume 1: Long Papers).

Demir, Ş. (2023). Neural Coreference Resolution for Turkish. *Journal of Intelligent Systems: Theory and Applications, 6*(1), 85-95.

Denis, P., & Baldridge, J. (2008). *Specialized models and ranking for coreference resolution.* Paper presented at the Proceedings of the Conference on Empirical Methods in Natural Language Processing.

Devlin, J., Chang, M.-W., Lee, K., & Toutanova, K. (2018). Bert: Pre-training of deep bidirectional transformers for language understanding. *arXiv preprint arXiv:.04805*.

Durrett, G., & Klein, D. (2014). A joint model for entity analysis: Coreference, typing, and linking. *Transactions of the association for computational linguistics*

*2*, 477-490.

Fallahi, F., & Shamsfard, M. (2011). Recognizing anaphora reference in Persian sentences. *International Journal of Computer Science Issues (IJCSI), 8*(2), 324.

Farahani, M., Gharachorloo, M., Farahani, M., & Manthouri, M. (2021). Parsbert: Transformer-based model for persian language understanding. *Neural Processing Letters, 53*, 3831-3847.

Graves, A., & Schmidhuber, J. (2005). Framewise phoneme classification with bidirectional LSTM and other neural network architectures. *Neural networks, 18*(5-6), 602-610.

Haji Mohammadi, H., Talebpour, A., Mahmoudi Aznaveh, A., & Yazdani, S. (2023). Mehr: A Persian Coreference Resolution Corpus. *Journal of AI and Data Mining*.

Hajimohammadi, H., Talebpour, A., & Mahmoudi Aznaveh, A. (2016). *Using decision tree to Persian pronoun resolution.* Paper presented at the Twenty-first annual National Conference of the Computer Society of Iran, Tehran.

Hirschman, L. (1997). MUC-7 coreference task definition, version 3.0. *Proceedings of MUC-7,*.

Hou, F., Wang, R., Ng, S.-K., Zhu, F., Witbrock, M., Cahan, S. F., . . . Jia, X. (2024). Anisotropic span embeddings and the negative impact of higher-order inference for coreference resolution: An empirical analysis. *Natural Language Engineering*, 1-22.

Joshi, M., Chen, D., Liu, Y., Weld, D. S., Zettlemoyer, L., & Levy, O. (2020). Spanbert: Improving pre-training by representing and predicting spans. *Transactions of the Association for Computational Linguistics, 8*, 64-77.

Joshi, M., Levy, O., Zettlemoyer, L., & Weld, D. S. (2019). *BERT for Coreference Resolution: Baselines and Analysis.* Paper presented at the Proceedings of the 2019 Conference on Empirical Methods in Natural Language Processing and the 9th International Joint Conference on Natural Language Processing (EMNLP-IJCNLP).

Kantor, B., & Globerson, A. (2019). *Coreference resolution with entity equalization.* Paper presented at the Proceedings of the 57th Annual Meeting of the Association for Computational Linguistics.

Kenton, J. D. M.-W. C., & Toutanova, L. K. (2019). *BERT: Pre-training of Deep Bidirectional Transformers for Language Understanding.* Paper presented at the Proceedings of NAACL-HLT.
Khosla, S., & Rose, C. (2020). *Using Type Information to Improve Entity Coreference Resolution.* Paper presented at the Proceedings of the First Workshop on Computational Approaches to Discourse.
Kingma, D. P., & Ba, J. (2014). Adam: A method for stochastic optimization. *arXiv preprint arXiv:1412.6980*.
Kriman, S., & Ji, H. (2021). *Joint detection and coreference resolution of entities and events with document-level context aggregation.* Paper presented at the Proceedings of the 59th Annual Meeting of the Association for Computational Linguistics and the 11th International Joint Conference on Natural Language Processing: Student Research Workshop.
Lai, T. M., Bui, T., & Kim, D. S. (2022). *End-to-end neural coreference resolution revisited: A simple yet effective baseline.* Paper presented at the ICASSP 2022-2022 IEEE International Conference on Acoustics, Speech and Signal Processing (ICASSP).
Lata, K., Singh, P., & Dutta, K. (2022). Mention detection in coreference resolution: survey. *Applied Intelligence*, 1-45.
Lee, H., Chang, A., Peirsman, Y., Chambers, N., Surdeanu, M., & Jurafsky, D. (2013). Deterministic coreference resolution based on entity-centric, precision-ranked rules. *Computational Linguistics, 39*(4), 885-916.
Lee, K., He, L., Lewis, M., & Zettlemoyer, L. (2017). *End-to-end Neural Coreference Resolution.* Paper presented at the Proceedings of the 2017 Conference on Empirical Methods in Natural Language Processing.
Lee, K., He, L., & Zettlemoyer, L. (2018). *Higher-Order Coreference Resolution with Coarse-to-Fine Inference.* Paper presented at the Proceedings of the 2018 Conference of the North American Chapter of the Association for Computational Linguistics: Human Language Technologies, Volume 2 (Short Papers).
Liu, R., Mao, R., Luu, A. T., & Cambria, E. (2023). A Brief Survey on Recent Advances in Coreference Resolution. *Artificial Intelligence Review*.
Liu, Z., Shi, K., & Chen, N. (2021). *Coreference-Aware Dialogue Summarization.* Paper presented at the Proceedings of the 22nd Annual Meeting of the Special Interest Group on Discourse and Dialogue.
Mao, R., & Li, X. (2021). *Bridging towers of multi-task learning with a gating mechanism for aspect-based sentiment analysis and sequential metaphor identification.* Paper presented at the Proceedings of the AAAI conference on artificial intelligence.
Mohammadi, H. H., Talebpour, A., Aznaveh, A. M., & Yazdani, S. (2022). A hybrid entity-centric approach to Persian pronoun resolution. *arXiv preprint arXiv:.06257*.
Moosavi, N. S., & Ghassem-Sani, G. (2009). A ranking approach to Persian pronoun resolution. *Advances in Computational Linguistics. Research in Computing Science, 41*, 169-180.
Murayshid, H. S., Benhidour, H., & Kerrache, S. (2023). A Sequence-to-Sequence Approach for Arabic Pronoun Resolution. *arXiv preprint arXiv:2305.11529*.
Nazaridoust, M., Bidgoli, B. M., & Nazaridoust, S. (2013). *Co-reference Resolution in Farsi Corpora.* Paper presented at the In Advance Trends in Soft Computing: Proceedings of WCSC 2013, Cham.
Nourbakhsh, A., & Bahrani, M. (2017). *Persian Pronoun Resolution Using Data Driven Approaches.* Paper presented at the International Conference on Information and Software Technologies.
Pradhan, S., Moschitti, A., Xue, N., Uryupina, O., & Zhang, Y. (2012). *CoNLL-2012 shared task: Modeling multilingual unrestricted coreference in OntoNotes.* Paper presented at the Joint Conference on EMNLP and CoNLL-Shared Task.
Rahimi, Z., & HosseinNejad, S. (2020). Corpus based coreference resolution for Farsi text *Signal and Data Processing*

*17*(1), 79-98.

Sahlani, H., Hourali, M., & Minaei-Bidgoli, B. (2020a). Coreference Resolution Using Semantic Features and Fully Connected Neural Network in the Persian Language. *International Journal of Computational Intelligence Systems*.

Sahlani, H., Hourali, M., & Minaei-Bidgoli, B. (2020b). Coreference resolution with deep learning in the persian language. *Signal and Data Processing*

Sil, A., Kundu, G., Florian, R., & Hamza, W. (2018). *Neural cross-lingual entity linking.* Paper presented at the Proceedings of the AAAI Conference on Artificial Intelligence.

Soon, W. M., Ng, H. T., & Lim, D. C. Y. J. C. l. (2001). A machine learning approach to coreference resolution of noun phrases. *Computational Linguistics, 27*(4), 521-544.

Stylianou, N., & Vlahavas, I. (2021). A neural entity coreference resolution review. *Expert Systems with Applications, 168*, 114466.

Sukthanker, R., Poria, S., Cambria, E., & Thirunavukarasu, R. (2020). Anaphora and coreference resolution: A review. *Information Fusion, 59*, 139-162.

Sutskever, I., Vinyals, O., & Le, Q. V. (2014). Sequence to sequence learning with neural networks. *Advances in neural information processing systems, 27*.

Vaswani, A., Shazeer, N., Parmar, N., Uszkoreit, J., Jones, L., Gomez, A. N., . . . Polosukhin, I. (2017). Attention is all you need. *Advances in neural information processing systems, 30*.

Wang, Y., Shen, Y., & Jin, H. (2021). *An End-To-End Actor-Critic-Based Neural Coreference Resolution System.* Paper presented at the ICASSP 2021-2021 IEEE International Conference on Acoustics, Speech and Signal Processing (ICASSP).

Wiseman, S., Rush, A. M., & Shieber, S. M. (2016). *Learning Global Features for Coreference Resolution.* Paper presented at the Proceedings of the 2016 Conference of the North American Chapter of the Association for Computational Linguistics: Human Language Technologies.

Wiseman, S. J., Rush, A. M., Shieber, S. M., & Weston, J. (2015). *Learning anaphoricity and antecedent ranking features for coreference resolution.* Paper presented at the Proceedings of the 53rd Annual Meeting of the Association for Computational Linguistics and the 7th International Joint Conference on Natural Language Processing.

Wu, W., Wang, F., Yuan, A., Wu, F., & Li, J. (2020). *CorefQA: Coreference resolution as query-based span prediction.* Paper presented at the Proceedings of the 58th Annual Meeting of the Association for Computational Linguistics.

Xia, P., Sedoc, J., & Van Durme, B. (2020). *Incremental Neural Coreference Resolution in Constant Memory.* Paper presented at the Proceedings of the 2020 Conference on Empirical Methods in Natural Language Processing (EMNLP).

Xu, L., & Choi, J. D. (2020). *Revealing the Myth of Higher-Order Inference in Coreference Resolution.* Paper presented at the Proceedings of the 2020 Conference on Empirical Methods in Natural Language Processing (EMNLP).

Yimam, S. M., Gurevych, I., de Castilho, R. E., & Biemann, C. (2013). *WebAnno: A flexible, web-based and visually supported system for distributed annotations.* Paper presented at the Proceedings of the 51st Annual Meeting of the Association for Computational Linguistics: System Demonstrations.

Yin, Q., Zhang, Y., Zhang, W., Liu, T., & Wang, W. Y. (2018). *Zero pronoun resolution with attention-based neural network.* Paper presented at the Proceedings of the 27th international conference on computational linguistics.

Zhang, R., dos Santos, C., Yasunaga, M., Xiang, B., & Radev, D. (2018). *Neural Coreference Resolution with Deep Biaffine Attention by Joint Mention Detection and Mention Clustering.* Paper presented at the Proceedings of the 56th Annual Meeting of the Association for Computational Linguistics (Volume 2: Short Papers).